\documentclass[10pt, a4paper]{article}
\usepackage[final]{lrec2026} 

\usepackage{booktabs} 

\usepackage{caption}
\usepackage{makecell}  
\usepackage{graphicx}  

\title{SynBullying: A Multi-LLM Synthetic Conversational Dataset for Cyberbullying Detection}

\name{Arefeh Kazemi$^{1}$,
Hamza Qadeer$^{1}$,
Joachim Wagner$^{1}$,
Hossein Hosseini$^{2}$,\\ {\bf \large Sri Balaaji Natarajan Kalaivendan$^{1}$,
Brian Davis$^{1}$}
}

\address{
$^{1}$School of Computing, ADAPT Centre, Dublin City University, Dublin, Ireland \\
$^{2}$University of Isfahan, Isfahan, Iran \\
\{arefeh.kazemi, hamza.qadeer, joachim.wagner, nk.sribalaaji, brian.davis\}@adaptcentre.ie \\
h.hosseini@eng.ui.ac.ir
}

\abstract{
We introduce SynBullying, a synthetic multi-LLM conversational dataset for studying and detecting cyberbullying (CB). SynBullying provides a scalable and ethically safe alternative to human data collection by leveraging large language models (LLMs) to simulate realistic bullying interactions. The dataset offers (i) conversational structure, capturing multi-turn exchanges rather than isolated posts; (ii) context-aware annotations, where harmfulness is assessed within the conversational flow considering context, intent, and discourse dynamics; and (iii) fine-grained labeling, covering various CB categories for detailed linguistic and behavioral analysis. We evaluate SynBullying across five dimensions, including conversational structure, lexical patterns, sentiment/toxicity, role dynamics, harm intensity, and CB-type distribution. We further examine its utility by testing its performance as standalone training data and as an augmentation source for CB classification.
 \\ \newline  }

\begin{document}

\maketitleabstract

\section{Introduction}
Cyberbullying (CB) is a pervasive form of online aggression that disproportionately affects children and adolescents \cite{hinduja2014bullying}. It is inherently interactional and context-dependent, emerging through multi-turn exchanges among bullies, victims, and bystanders \cite{sheth2022defining}. While context is central to most abusive language detection tasks, CB presents unique challenges because the harmfulness of a message often depends on conversational history, relational cues, and patterns of escalation, rather than being apparent from a single message.\cite{ziems2020aggressive}.
Numerous datasets have been developed for the detection of CB and toxicity \cite{van2018automatic, chatzakou2017mean, wulczyn2017ex}. Although these corpora have advanced abuse and harassment classification, most are post-level or message-level datasets, where annotations are independent of conversational context \cite{ziems2020aggressive}. Consequently, models trained on them often fail to capture the interactional intent, indirect aggression, and the discourse framing that define bullying discourse. Moreover, adolescent online communication is linguistically dynamic, shaped by evolving slang, emojis, multimodal references, and coded expressions \cite{dembe2024impact, mcgillivray2022leveraging}. Models trained on static data struggle to generalize to emerging or obfuscated bullying patterns.  
Collecting authentic conversational data from minors is constrained by ethical, legal, and psychological considerations \cite{facca2020exploring}. Annotation of harmful content is resource-intensive and ethically sensitive, potentially exposing annotators to distress and re-traumatization \cite{alemadi2024emotional}. These barriers have created a data scarcity problem, limiting progress in context-aware CB detection and computational modeling of peer aggression.
To address these limitations, we introduce SynBullying, a synthetic multi-LLM conversational dataset for studying and detecting CB.\footnote{Publicly available
    at \url{https://huggingface.co/datasets/arrkaa-NLP/SynBullying}.
}
SynBullying provides a viable, scalable and ethically safe alternative to human data collection by leveraging large language models (LLMs) to simulate realistic bullying interactions. The dataset offers (i) conversational structure, capturing multi-turn exchanges rather than isolated posts; (ii) context-aware annotations, where harmfulness is assessed within the conversational flow considering context, intent, and discourse dynamics; and (iii) fine-grained labeling, covering various CB categories for detailed linguistic and behavioral analysis. We evaluate SynBullying across six dimensions and examine its utility by testing its performance as standalone training data and as an augmentation source for CB classification.

\section{Related Work}
CB is a complex form of online aggression characterized by intentionality, power imbalance, and multi-turn conversational dynamics \cite{patchin2006bullies, van2018automatic, ziems2020aggressive, emmery2021current}. CB exhibits relational and behavioral patterns such as exclusion, denigration, flaming, and harassment, as well as nuanced bystander behaviors including enabling, defending, or passive observation \cite{nadali2013review, slonje2013nature, bauman2015types, leung2018you, song2018factors, ollagnier-etal-2022-cyberagressionado}. Fine-grained, role-aware datasets \cite{van2018automatic, sprugnoli-etal-2018-creating, ollagnier-etal-2022-cyberagressionado} have advanced CB research, but remain limited in scale, class balance, ecological validity, and youth representativeness.
Recent advances in large language models (LLMs) have enabled scalable generation of high-quality synthetic data, offering controllable and ethically safer alternatives to human-annotated datasets. Early work leveraged LLMs for data augmentation, knowledge distillation, and few-shot learning \cite{anaby-etal-2020-donot, he-etal-2021-generate, he-etal-2022-generate, bonifacio-etal-2022-inpars, meng-etal-2022-generating, yoo-etal-2021-gpt3mix}, while domain-specific studies generated synthetic medical dialogues \cite{wang-etal-2024-notechat} and socially-aware datasets for suicidal ideation detection \cite{ghanadian-etal-2024-socially}. In online harm research, LLM-based augmentation has been applied to toxic language detection \cite{schmidhuber-kruschwitz-2024-llm}, harmful and biased content \cite{electronics13173431, hui-etal-2024-toxicraft}, and recently to CB detection \cite{kazemi2025synthetic, tari2025highfidelitysyntheticmultiplatformsocial}. Other studies have generated CB data via simulations or structured models, including Bayesian networks for serious game scenarios \cite{perez-etal-2024-generation} or hybrid synthetic-authentic datasets \cite{ejaz-etal-2024-multi}. These approaches illustrate the potential of LLMs for high-coverage datasets, though many remain limited to single-turn messages, small scales, or reliance on seed data, without fully modeling multi-turn conversational dynamics. While prior work has explored role-play corpora and LLM-based approaches for CB, there remains limited research on large-scale, multi-turn synthetic CB datasets for adolescents that integrate context-aware annotations, fine-grained categorization, and systematic validation. 

\section{Dataset Construction}
\subsection{Authentic Dataset}\label{s:authentic}

To evaluate the realism of our synthetic CB data, we benchmark it against an authentic CB dataset created through teen role-play sessions \cite{sprugnoli-etal-2018-creating}. We use the English version of this dataset \cite{verma-etal-2023-leveraging}. Each simulated conversation assigns participants to predefined roles that mirror real-world CB dynamics: \textit{Victim}, \textit{Bully}, \textit{Bully Supporter}, and \textit{Victim Supporter}. This structured setup enables the emergence of socially meaningful power hierarchies commonly observed in CB interactions. All conversations are initiated using one of four pre-designed CB-triggering scenarios (labeled A–D), each describing a realistic situation that could lead to peer conflict and harassment. 
All messages in the dataset were manually annotated by expert annotators using the fine-grained CB types introduced by
\newcite{van-hee-et-al-2015-guidelines}\footnote{Since
     we only annotate harmful messages with a 
     CB types in this paper, the ``Defense'' category is
    excluded.}: 
\textit{Threat or Blackmail}, \textit{Insult General}, \textit{Insult Body Shame}, \textit{Insult Discrimination Sexism}, \textit{Insult Discrimination Racism}, 
\textit{Insult Attacking Relatives}, \textit{Curse or Exclusion}, \textit{Defamation}, 
\textit{Sexual Talk Harmless}, \textit{Sexual Talk Harassment}, \textit{Encouragement to Harassment}, \textit{Other}.

\subsection{Synthetic Datasets}
We generate synthetic data using three LLMs: GPT-4o (Feb-2025 version)~\cite{openai-2024-gpt}, Llama-3.3-70B-Instruct~\cite{meta-2024-llama33}, and Grok-2 (Feb-2025 version)~\cite{xai-2024-grok}. All synthetic data are labeled using GPT-4o (Sept-2025 version). Our prompt engineering process begins with a basic template and evolves through iterative refinement. In each iteration, we adjust the prompt based on the qualitative evaluation of the LLM outputs on a development set, improving its ability to consistently produce relevant and high-quality conversations and labels.

\subsubsection{Synthetic Data Generation}

We target the generation of multi-turn conversations that explicitly contain CB. To this end, we define a role-based prompt in which \textbf{eleven} fictional teenage participants are assigned specific roles: one victim, two bullies, four victim supporters, and four bully supporters. To encourage the model to produce harmful content within a safe and research-driven setting, we frame the task as academic work conducted for CB detection. The model is then provided with a predefined CB scenario and is instructed to generate realistic, profanity-rich CB conversations. Models occasionally refuse to produce harmful content; in such cases, we reissue the prompt until the desired number of conversations is obtained. To ensure consistency with the authentic dataset, we embed the original role-play scenarios (A–D) from \newcite{sprugnoli-etal-2018-creating}, which guide the narrative progression and participant interactions.\footnote{The final version of this paper will show the prompts used for synthetic data generation in an appendix.} Finally, the LLMs generate cyberbullying (CB) conversations as sequences of ordered messages exchanged among participants. Each message is assigned a specific role (Victim, Bully 1, Bully 2, Victim Supporter 1–4, or Bully Supporter 1–4) reflecting the social dynamics of the interaction. Collectively, the messages in each conversation depict a coherent CB incident occurring between these roles. 

\subsubsection{Synthetic Labeling}
Following data generation, we automatically annotate with LLMs each message with a binary harmful/harmless label.  Harmful messages are further annotated with one or more CB type labels. Since CB classification is context-aware, we perform annotation on full conversations rather than individual isolated messages. Each LLM call, therefore, processes an entire conversation and returns labels for all messages. Based on our initial evaluations, GPT-4o demonstrates the highest performance among tested models for CB-related labeling tasks; therefore, we employ it for all synthetic annotations. Each message receives two labels: (1) is\_harmful = yes/no, and (2) if applicable, a set of CB-type labels aligned with the taxonomy used in the authentic dataset described in Section \ref{s:authentic}. Note that each harmful message can be assigned to more than one CB type.

\section{Label Quality}\label{sec:label_quality}
We evaluate the reliability of GPT-4o as an annotator on the authentic dataset by comparing its predictions with human gold labels for both binary harmfulness and fine-grained CB type classification. This analysis informs the reliability of GPT-4o labeling for our synthetic conversational dataset, \textbf{SynBullying}.

\subsection{Binary Label Evaluation}
Table \ref{t:confusion-matrix} and Table~\ref{t:gpt4o_human_agreement_binary} report the confusion matrix, inter-annotator agreement, and performance metrics for the \textit{harmful} versus \textit{harmless} task, respectively. The confusion matrix shows a balanced performance of GPT-4o in labeling messages, with a slight tendency to over-predict the \textit{harmful} class. The model correctly identifies most harmful messages but occasionally misclassifies borderline, sarcastic, or context-dependent instances.Cohen’s $\kappa = 0.627$ and Fleiss’ $\kappa = 0.625$ indicate substantial agreement~\cite{landis1977measurement}. The human–human Cohen’s $\kappa$ for this task is $0.69$
\cite{van-hee-et-al-2015-guidelines},
also substantial. These results show that GPT-4o can approximate human-level reliability for binary harmful content detection on the authentic dataset.  The combination of high recall ($0.825$) and slightly lower precision ($0.688$) suggests GPT-4o is sensitive to harmful content, prioritizing coverage over minimizing false positives, which is desirable in safety-critical NLP applications. 

\begin{table}[h!]
\centering
\begin{tabular}{c|cc}
\hline
Human label & \multicolumn{2}{c}{LLM prediction} \\
\cline{2-3}
 & 0 & 1 \\
\hline
0 & 1279 & 249 \\
1 & 116 & 548 \\
\hline
\end{tabular}
\caption{Confusion matrix comparing GPT-4o labels with human gold labels for binary CB classification on the authentic dataset}
\label{t:confusion-matrix}
\end{table}

\begin{table}[htbp]
\centering
\begin{tabular}{l c |c c}
\toprule
Metric & Score &Metric & Score \\
\midrule
Cohen’s K   & 0.627 & Fleiss’ K   & 0.625 \\
Accuracy        & 0.833 & Precision       & 0.688 \\
Recall          & 0.825 & F1 Score        & 0.750 \\
\bottomrule
\end{tabular}
\caption{Inter-annotator Agreement and Performance Metrics for GPT-4o Labels Vs Human Gold Labels for binary CB classification on the Authentic Dataset}
\label{t:gpt4o_human_agreement_binary}
\end{table}

\subsection{CB Type Evaluation}

\begin{figure*}[!ht]
\begin{center}
\includegraphics[width=\textwidth]{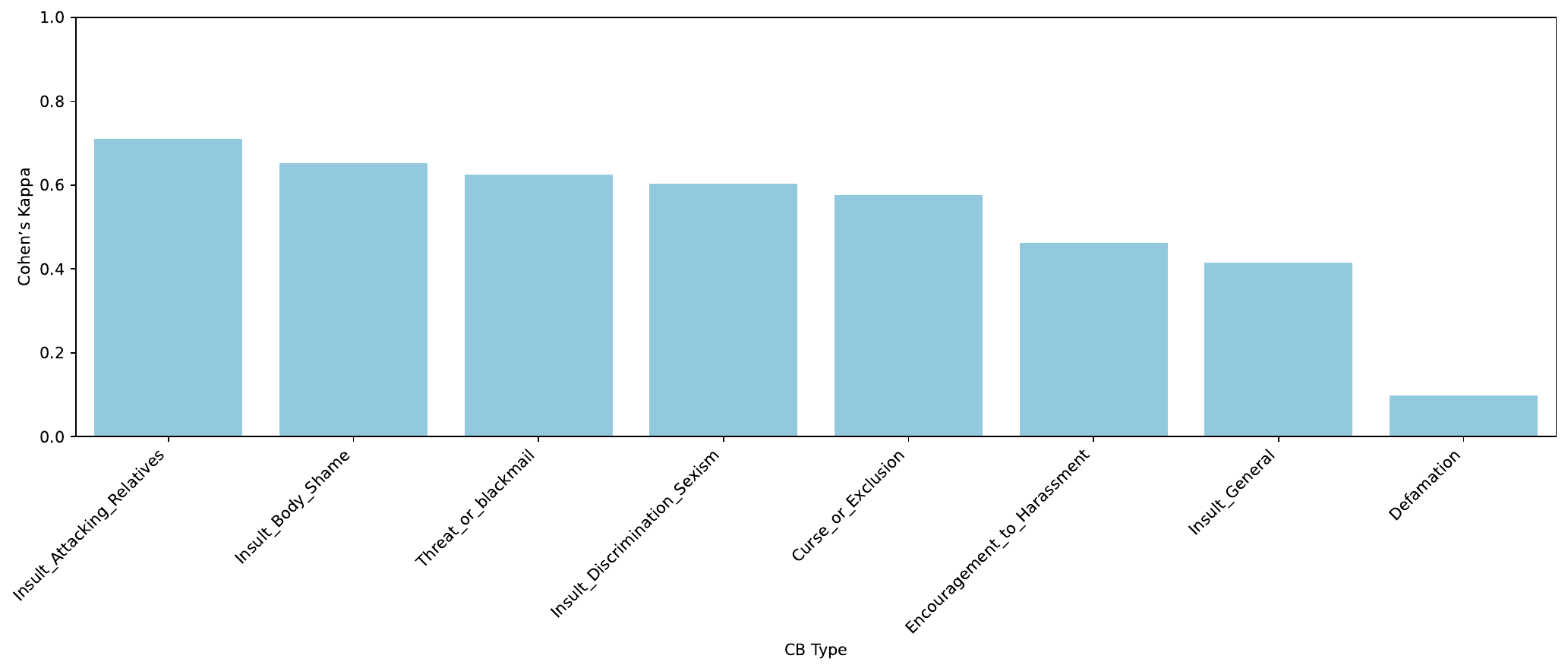}
\caption{Inter-annotator agreement between GPT-4o and human gold labels for CB type classification on the authentic dataset}
\label{f:cohen-kappa}
\end{center}
\end{figure*}

\begin{figure*}[!ht]
\begin{center}
\includegraphics[width=\textwidth]{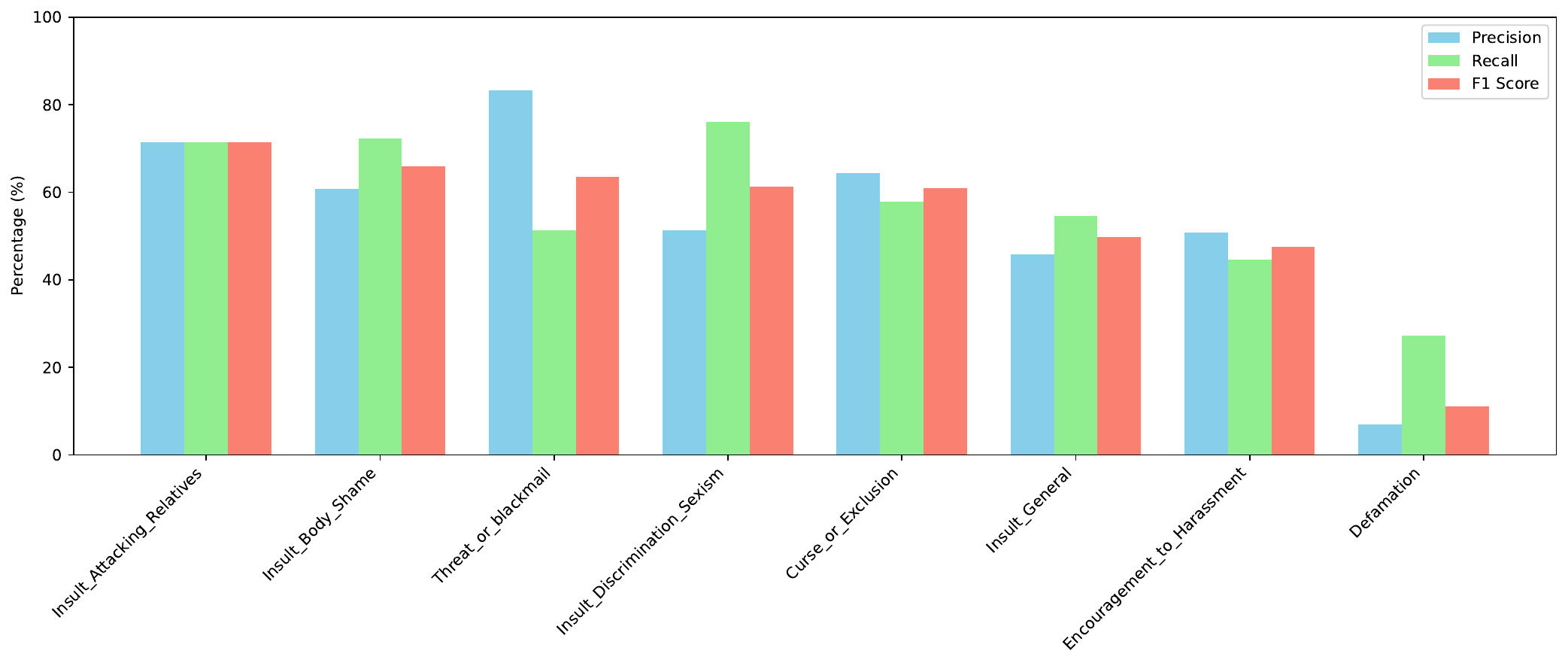}
\caption{Performance metrics of GPT-4o for predicting CB types, evaluated against human gold labels on the authentic dataset.}
\label{f:performance-binary-label}
\end{center}
\end{figure*}

We next assess GPT-4o’s labeling quality for fine-grained CB type classification. Figure~\ref{f:cohen-kappa} presents the inter-annotator agreement between GPT-4o and human gold labels across CB categories. Performance metrics per CB type are shown in Figure~\ref{f:performance-binary-label}. GPT-4o performs best on explicit categories such as \textit{Insult\_Attacking\_Relatives} ($F_{1}=71.43$) and \textit{Insult\_Body\_Shame} ($F_{1}=66.02$), which contain direct and lexically salient insults. Moderate performance is observed for \textit{Threat\_or\_Blackmail} ($F_{1}=63.49$) and \textit{Curse\_or\_Exclusion} ($F_{1}=60.96$), where the model must integrate pragmatic or modal cues. More context-dependent types such as \textit{Encouragement\_to\_Harassment} ($F_{1}=47.54$) and \textit{Defamation} ($F_{1}=11.11$) remain challenging, often requiring understanding of implicit or reputational harm. Notably, although F1 scores and agreement for categories such as \textit{Defamation} are very low, these metrics align with human–human agreement
($\kappa = 0.0$)
\cite{van-hee-et-al-2015-guidelines}
This indicates that these CB types are inherently subjective and context-dependent, making consistent labeling difficult even for humans~\cite{xu2012learning}. Therefore, GPT-4o’s performance on these categories reflects intrinsic annotation difficulty rather than model deficiency. 

\subsection{Comparison with Human--Human Agreement}

\begin{figure*}[!ht]
\begin{center}
\includegraphics[width=\textwidth]{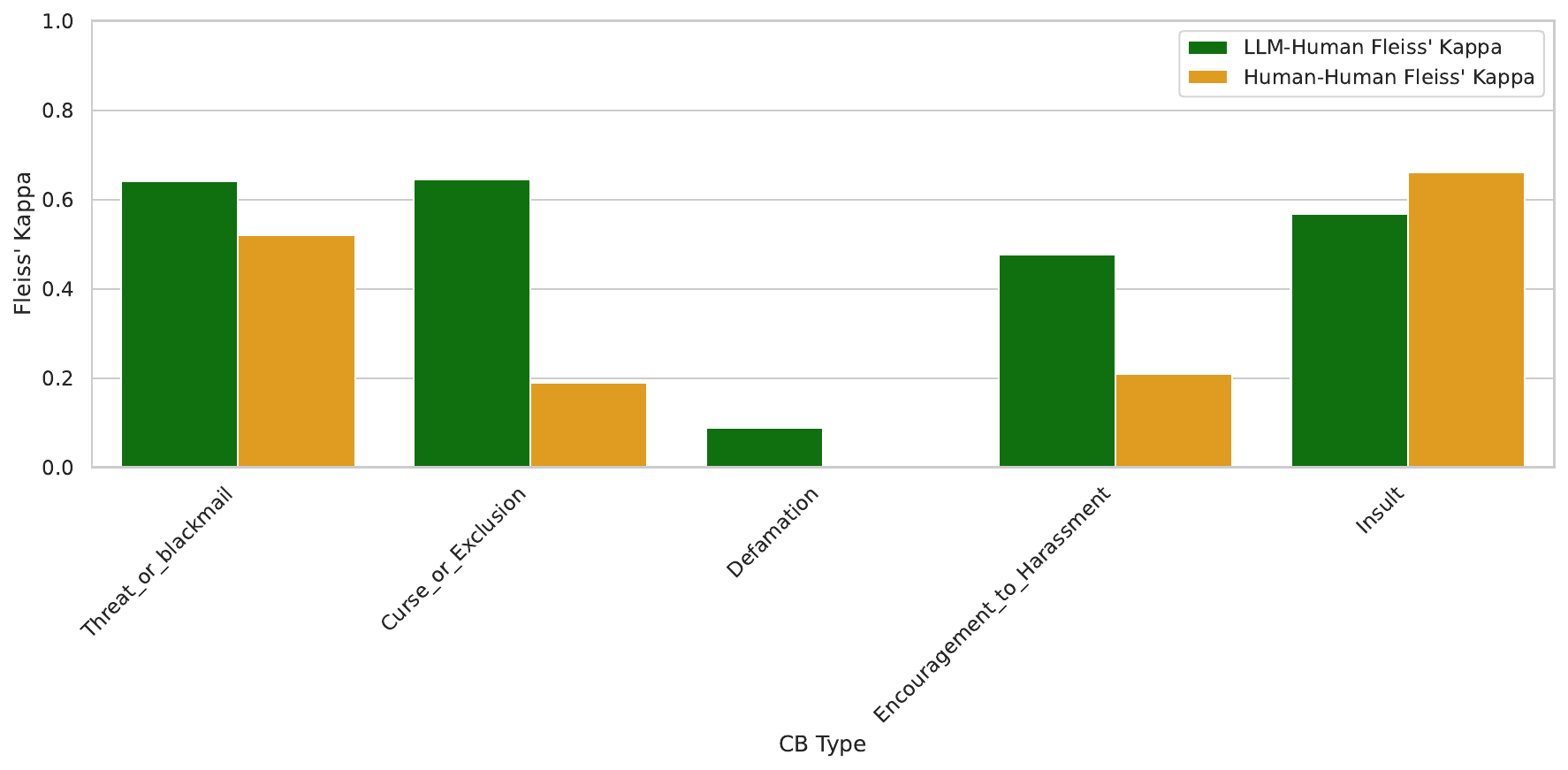}
\caption{Fleiss’ Kappa scores for each CB type, with all insult-related types aggregated. The figure compares inter-annotator agreement between GPT-4o and human gold labels with the agreement among human annotators.}
\label{f:fleiss-kappa}
\end{center}
\end{figure*}

Figure~\ref{f:fleiss-kappa} compares Fleiss’~$\kappa$ scores for GPT-4o--human versus human--human agreement. Human annotators achieved $\kappa$ values ranging from $0.00$ (slight) for \textit{Defamation} to $0.66$ (substantial) for \textit{Insult}, demonstrating the inherent difficulty of subjective CB type labeling. GPT-4o reached comparable or higher reliability in several categories: \textit{Threat\_or\_Blackmail}, \textit{Curse\_or\_Exclusion}, and \textit{Encouragement\_to\_Harassment}.
These results demonstrate that GPT-4o can reproduce human-level annotation quality and, in some ambiguous cases, even surpass inter-human agreement. Comparable or even higher levels of consistency in LLM-generated annotations have also been reported previously in other domains~\citep{bojic2025comparing}. Nonetheless, subtle CB types that require pragmatic understanding remain challenging for both LLMs and humans to annotate, highlighting the intrinsic difficulty of CB type detection. Given this demonstrated reliability, GPT-4o was adopted as the primary annotator for our synthetic dataset, SynBullying, providing strong justification for scaling labeling to synthetic conversational data while maintaining comparable label quality and consistency with human-labeled data.

\section{Comparative Analysis of Datasets}
We evaluate the linguistic and behavioral realism of the synthetic datasets in comparison to authentic CB conversations across five dimensions. We then assess their practical utility for CB detection by examining classifier performance when trained solely on synthetic data and when synthetic data is used to augment authentic training data.

\subsection{Lexical Diversity and Conversational Statistics}

\begin{table*}[htbp]
\centering
\begin{tabular}{lcccccc}
\toprule
Dataset & \#Conv & \#Messages & Avg. Msg/Conv & Avg. Tokens/Msg & MTLD & Vocab Size \\
\midrule
Authentic & 10 & 2192 & 219.20 & 8.34 & 56.27 & 2665 \\
GPT-4o & 40 & 4770 & 119.25 & 9.30 & 75.54 & 3316 \\
Grok & 40 & 3960 & 99.00 & 16.51 & 51.69 & 2844 \\
LLaMA & 40 & 3300 & 82.50 & 15.79 & 51.55 & 2962 \\
\bottomrule
\end{tabular}
\caption{Conversational and lexical metrics. MTLD = Measure of Textual Lexical Diversity.}
\label{tab:linguistic_metrics}
\end{table*}

Table \ref{tab:linguistic_metrics} shows the conversational structure and lexical characteristics of the datasets. We use the Measure of Textual Lexical Diversity (MTLD) to evaluate the lexical variation of synthetic and authentic datasets, as it offers a length-independent metric of lexical diversity \cite{mccarthy2010mtld}.
Considering intrinsic lexical and conversational metrics, LLaMA most closely resembles the authentic dataset. Its MTLD and average tokens per message are nearer to the authentic values than Grok, suggesting that LLaMA better preserves the lexical richness and typical message length patterns of human conversations. While Grok is slightly closer to the authentic dataset in terms of raw vocabulary size, LLaMA provides a more balanced match across multiple linguistic dimensions. GPT-4o, despite having the highest MTLD and largest vocabulary, deviates substantially from authentic norms, reflecting highly stylized and over-diverse outputs. Overall, in terms of intrinsic lexical and conversational realism, LLaMA most closely resembles the authentic dataset, followed by Grok and then GPT-4o, indicating that LLaMA provides the most faithful replication of natural linguistic CB patterns among the synthetic datasets.

\subsection{Sentiment, Toxicity, and Offensive Language Analysis}

\begin{table*}[htbp]
\centering
\begin{tabular}{lccccc}
\toprule
Dataset & Profanity (/100 tokens) & Toxicity (\%) & Positive (\%) & Neutral (\%) & Negative (\%) \\
\midrule
Authentic & 0.78 & 19.21 & 30.38 & 44.75 & 24.86 \\
GPT-4o & 0.03 & 1.95 & 48.72 & 33.88 & 17.40 \\
Grok & 1.36 & 39.80 & 39.24 & 14.87 & 45.88 \\
LLaMA & 0.26 & 15.42 & 49.70 & 17.61 & 32.70 \\
\bottomrule
\end{tabular}
\caption{Sentiment distribution, profanity, and toxicity rates.}
\label{tab:sentiment_toxicity}
\end{table*}

Table \ref{tab:sentiment_toxicity} shows the sentiment distribution, profanity, and toxicity rates across datasets. Sentiment was measured using NLTK's \cite{loper-bird-2002-nltk} VADER analyzer \cite{hutto2014vader} with standard compound score thresholds ($\geq 0.05 =$ positive, $\leq -0.05 =$ negative). Profanity detection used a curated lexicon with pattern matching for censored variants (e.g., f*k, sht). Toxicity was scored using ToxicBERT \cite{unitary2020toxicbert}, with messages scoring $\geq 0.5 =$ flagged as toxic. All metrics were normalized per 100 tokens after custom tokenization that preserves censored forms and handles elongations. Table \ref{tab:sentiment_toxicity} shows the authentic dataset contains moderate levels of toxicity (19.21\%) and profanity (0.78), reflecting realistic online hostility. In contrast, GPT-4o exhibits a strong safety alignment, with drastically reduced toxicity (1.95\%) and profanity (0.03), alongside inflated positive sentiment and fewer negative messages. Grok diverges most significantly, producing the highest toxicity (39.80\%) and profanity (1.36\%), heavily skewing toward negative sentiment and substantially reducing neutral content. LLaMA occupies a middle ground, showing moderate toxicity (15.42\%) and sentiment distributions closer to the authentic dataset. Overall, LLaMA best approximates the emotional and toxicity profile of authentic conversations, GPT-4o overly sanitizes discourse, and Grok exaggerates aggression beyond realistic levels.

\subsection{Role-Based Interaction Patterns}

\begin{table}[htbp]
\centering
\begin{tabular}{lcccc}
\toprule
Role & Authentic & GPT-4o  & Grok  & LLaMA  \\
\midrule
V & 16.01 & 14.65 & 13.81 & 19.58 \\
VS & 28.65 & 43.40 & 32.25 & 31.12 \\
B & 28.74 & 18.22 & 27.78 & 26.18 \\
BS & 26.28 & 23.73 & 26.16 & 23.12 \\
\bottomrule
\end{tabular}
\caption{Percentage (\%) of messages by main CB roles: V (Victim), B (Bully), VS (VictimSupport), BS (BullySupport)}
\label{tab:role_distribution}
\end{table}

Table~\ref{tab:role_distribution} illustrates the percentage of messages contributed by each main CB role across datasets. In the authentic dataset, bully messages clearly exceed victim messages, while support roles (VictimSupport + BullySupport) contribute 54.93\%, reflecting typical power dynamics in cyberbullying interactions. Among synthetic datasets, Grok closely mirrors this distribution, preserving authentic conversational hierarchies. LLaMA also maintains the bully–victim imbalance but slightly amplifies victim contributions. In contrast, GPT-4o deviates from this pattern, equalizing bully and victim participation and amplifying VictimSupport, consistent with a safety-oriented design that reduces aggressive interactions.
\subsection{Harm Intensity}

\begin{table}
\centering
\begin{tabular}{lcc}
\hline
\textbf{Dataset} & \textbf{Harmful (\%)} & \textbf{Harmless (\%)} \\
\hline
Authentic & 36.36 & 63.64 \\
GPT       & 34.53 & 65.47 \\
Grok      & 53.56 & 46.44 \\
LLaMA     & 38.64 & 61.36 \\
\hline
\end{tabular}
\caption{Percentage of harmful and harmless messages.}
\label{t:harmful-harmless}
\end{table}

\begin{table}[htbp]
\centering
\begin{tabular}{lcc}
\toprule
Comparison & JSD & p-value \\
\midrule
Authentic vs GPT-4o & 0.0002 & 0.144 \\
Authentic vs Grok & 0.0150 & 3.81e-38 \\
Authentic vs LLaMA & 0.0003 & 0.0937 \\
\bottomrule
\end{tabular}
\caption{JSD and p-values for harmful vs. harmless message distributions. Lower JSD indicates higher similarity.}
\label{tab:jsd_harm}
\end{table}

Table~\ref{t:harmful-harmless} shows the percentage of harmful and harmless messages for each dataset. Table~\ref{tab:jsd_harm} reports the Jensen–Shannon divergence (JSD) \cite{sason2022divergence} and associated p-values for the marginal distribution of harmful vs. harmless messages. Both GPT-4o and LLaMA closely replicate the authentic dataset’s balance between harmful and harmless messages, showing very similar percentages of harmful messages and extremely low Jensen–Shannon divergence, indicating high alignment with authentic data. In contrast, Grok overestimates harmful content, resulting in a noticeably higher proportion of harmful messages and a larger divergence from the authentic distribution. Overall, GPT-4o and LLaMA exhibit strong fidelity to authentic harm patterns, while Grok produces a more aggressive synthetic dataset.

\subsection{CB Type Distributions}

\begin{figure*}[!ht]
\begin{center}
\includegraphics[width=\textwidth]{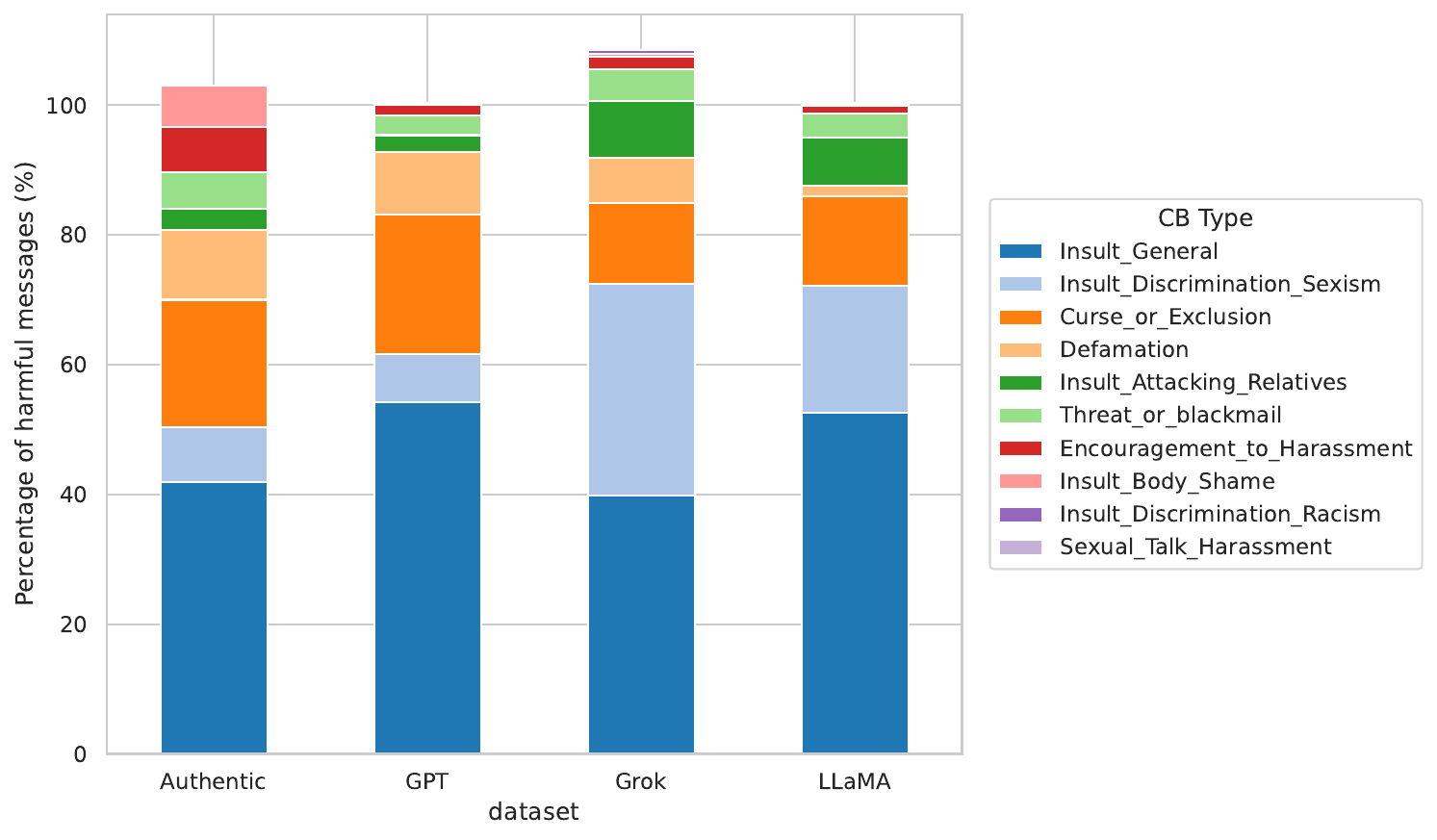}
\caption{Distribution of CB types}
\label{f:cb-types}
\end{center}
\end{figure*}

Figure~\ref{f:cb-types} presents the distribution of CB types across datasets\footnote{Since a single message may correspond to multiple CB types, the sum of percentages can exceed 100\%.}. Table~\ref{tab:jsd_cb} reports JSD and p-values for CB-type distributions. \textit{Insult\_General} is the most frequent type in all datasets. Secondary CB types vary across LLMs. GPT-4o best preserves secondary types such as \textit{Curse\_or\_Exclusion} and \textit{Defamation}, whereas Grok and LLaMA overrepresent discrimination-related insults. GPT-4o exhibits the closest match to the authentic CB-type distribution (JSD = 0.0317), while Grok and LLaMA deviate more. Although GPT-4o is closest, all synthetic CB-type distributions are significantly different from authentic data (p $<$ 1e-28), reflecting model-specific biases in generating harmful content.

\begin{table}[htbp]
\centering
\begin{tabular}{lcc}
\toprule
Comparison & JSD & p-value \\
\midrule
Authentic vs GPT-4o & 0.0317 & 1.11e-28 \\
Authentic vs Grok & 0.0739 & 4.33e-68 \\
Authentic vs LLaMA & 0.0803 & 7.85e-59 \\
\bottomrule
\end{tabular}
\caption{JSD and p-values for CB-type distributions. Lower JSD indicates higher similarity.}
\label{tab:jsd_cb}
\end{table}

\subsection{Classification Experiments}
\subsubsection{Experiments}
For harm detection, we adapt the BERT-base-uncased architecture with 110M parameters \cite{devlin-etal-2019-bert} by attaching a simple linear layer for binary classification over message-level inputs. To ensure robustness against training randomness, we repeat each experiment at least ten times with different random initialization and report average scores across runs. We conducted
\textbf{four}
groups of experiments to evaluate how synthetic data can support or challenge CB detection. \textbf{Experiment Group 0} serves as the authentic-only baseline, where models are trained and tested on the authentic dataset. \textbf{Experiment Group 1} examines the use of synthetic data instead of authentic data for training CB classifier. \textbf{Experiment Group 2 }reverses this direction to evaluate the robustness of CB classifiers trained solely on authentic data when exposed to LLM-generated harmful content.
\textbf{Experiment Group 3} 
explores
data augmentation, combining synthetic and authentic data to test whether synthetic examples improve detection of real 
harmful content. Given the small size of the authentic dataset (10 conversations), we apply leave-one-conversation-out cross-validation to ensure robust use of all data.
Although the authentic and synthetic sets differ in scale and style, this setup provides a fair and transparent basis for comparing domain transfer and augmentation effects.

\subsubsection{Results}
Table \ref{tab:results} shows the performance of a BERT-based classifier trained on authentic (Auth) and synthetic conversations. The authentic-only baseline (Experiment 0) achieves 68.2\% Harm-F1 and 80.7\% accuracy. Training solely on synthetic data (Exp 1) transfers poorly to authentic conversations, with LLaMA performing best, Grok moderate, and GPT worst. Reverse transfer (Experiment 2), where authentic-trained models are tested on synthetic content, shows strong detection for Grok-synthetic messages (75.7\% Hram-F1), moderate for LLaMA (61.7\%), and very low for GPT (37.9\%), highlighting blind spots in authentic-only classifiers. Augmenting authentic data with synthetic examples (Experminent 3) restores near-baseline performance on Auth (Auth+LLaMA: 68.7\% Harm-F1).
Overall, LLaMA is most effective for realistic augmentation, Grok for adversarial worst-case evaluation, and GPT-4o for exposing system blind spots, demonstrating the complementary value of combining synthetic and authentic datasets for robust moderation.

\begin{table}[t]
\centering
\small
\begin{tabular}{@{}clcc@{}}
\toprule
Exp & Train $\rightarrow$ Test & Harm-F1 & Acc. \\ \midrule
0 & Auth $\rightarrow$ Auth & 68.2 & 80.7 \\
\hline
1 & GPT $\rightarrow$ Auth & 49.0 & 53.7 \\
1 & Grok $\rightarrow$ Auth & 50.5 & 57.7 \\
1 & LLaMA $\rightarrow$ Auth & 53.4 & 71.9 \\
\hline
2 & Auth $\rightarrow$ GPT & 37.9 & 68.4 \\
2 & Auth $\rightarrow$ Grok & 75.7 & 74.8 \\
2 & Auth $\rightarrow$ LLaMA & 61.7 & 71.5 \\
\hline
3 & Auth+GPT $\rightarrow$ Auth & 65.8 & 78.1 \\
3 & Auth+Grok $\rightarrow$ Auth & 67.7 & 80.6 \\
3 & Auth+LLaMA $\rightarrow$ Auth & 68.7 & 81.2 \\
\bottomrule
\end{tabular}
\caption{Performance of a BERT-based classifier trained on authentic (Auth) and synthetic conversations generated by GPT, Grok, and LLaMA. Results report F1-score on the harmful class and overall accuracy (\%). Experiments cover baseline (Exp 0), domain transfer (Exp 1–2), and data augmentation (Exp 3–4).}
\label{tab:results}
\end{table}

\section{Discussion}

Table~\ref{tab:synthetic_ranking} summarizes the similarity of each synthetic dataset to the authentic WA dataset across key metrics. LLaMA generates messages that closely match authentic lexical variety, sentiment/toxicity patterns, role distributions, and harmful/harmless balance, making it ideal for realistic modeling and augmentation. Grok produces the highest levels of toxicity and profanity, with a strong bias toward negative sentiment and harmful messages. This makes it particularly useful for stress-testing moderation systems under realistic worst-case scenarios. GPT-4o generates lexically diverse, safer outputs while preserving harmful message proportions, making it particularly effective for exposing blind spots in authentic-only classifiers. The classifier results align with these qualitative properties. Models trained solely on synthetic data transfer poorly to authentic conversations, with LLaMA performing best, Grok moderate, and GPT worst. Reverse transfer shows Grok is easiest to detect, LLaMA moderately detectable, and GPT largely undetected, reflecting divergence from authentic conversations. Augmenting authentic data with synthetic examples restores or slightly improves detection on WA and enables robust detection of LLM-generated content. Detecting both human- and LLM-generated harmful content is essential, as LLMs can be used by humans or bots to generate cyberbullying in social media. Overall, training solely on synthetic data cannot replace authentic human-annotated data, as it does not fully capture the distribution of real-world harmful content. However, combining synthetic and authentic datasets provides a scalable strategy to improve detection, evaluate vulnerabilities, and ensure moderation systems generalize to diverse harmful content. 

\begin{table}[htbp]
\centering
\begin{tabular}{lccc}
\toprule
Metric & GPT & Grok & LLaMA \\
\midrule
Lexical \& Conv.       & *   & **  & *** \\
Sentiment/Toxicity     & *   & **  & *** \\
Role Distribution      & *   & *** & *** \\
Harm Intensity         & *** & *   & *** \\
CB-type Distribution   & *** & **  & *   \\
\bottomrule
\end{tabular}
\caption{Ranking of synthetic datasets based on similarity to the authentic dataset across five metrics. More stars indicate higher similarity.}  
\label{tab:synthetic_ranking}
\end{table}

\section{Conclusion and Future Work}


This paper introduced \textbf{SynBullying}, a multi-LLM conversational dataset for studying and detecting cyberbullying. Through systematic comparison with an authentic teen role-play corpus, we showed that synthetic data can approximate human conversational dynamics across multiple dimensions of realism. Among models,
LLaMA generated the most balanced and authentic-like interactions,
Grok produced extremely toxic examples useful for stress testing,
and GPT-4o provided lexically rich but safety-aligned outputs that expose model blind spots.
Empirical data suggests
that synthetic data alone cannot substitute for authentic annotations.
These findings highlight the complementary value of synthetic and authentic resources for building scalable, viable and ethically responsible CB detection systems. 

Future work will focus on expanding SynBullying to multilingual settings, enabling research on CB detection in low-resource languages and cross-lingual contexts. We also plan to investigate cross-platform generalization by modeling social context, including platform-specific interaction patterns and user role dynamics, to better capture the diversity of real-world online interactions. Insights from our evaluation of synthetic datasets will guide iterative improvements in prompt engineering to produce higher-quality, contextually coherent, and socially plausible conversations. Moreover, selected samples of the synthetic data will be validated with social scientists to ensure that the generated interactions are representative of authentic CB behaviors and reflect realistic social dynamics. These combined efforts aim to enhance both the reliability of synthetic data for model training and its value for studying the linguistic, behavioral, and social aspects of CB in diverse settings.

\section{Ethical Considerations}


The creation and use of datasets in the cyberbullying domain inherently involve sensitive content, including offensive language, threats, and harassment. Both authentic and synthetic datasets carry ethical risks, particularly if accessed or misused by unauthorized individuals. While synthetic datasets mitigate some privacy concerns by avoiding direct use of real user data, they can still reproduce harmful patterns, biases, or unrealistic scenarios that could influence downstream models in undesirable ways.

\paragraph{Privacy and Anonymity}  
All authentic data used in this study were anonymized to remove personally identifiable information. Synthetic datasets were generated without any access to private user information, ensuring that no sensitive personal data is exposed.

\paragraph{Bias and Representational Concerns}  
Synthetic datasets can inadvertently amplify model biases or skew the representation of certain behaviors. Our analysis shows that models differ in how they reproduce harmfulness intensity, sentiment, role distribution, and CB-type allocation. For instance, GPT-4o generates safer, support-oriented conversations, potentially underrepresenting aggressive interactions, whereas Grok tends to overrepresent hostile content. Recognizing these biases is crucial to prevent misleading conclusions or unsafe model behavior in downstream applications.

\paragraph{Usage Restrictions}  
Access to these datasets should be limited to researchers and practitioners working on cyberbullying detection, moderation, or related NLP tasks. The datasets should not be used to generate harmful content or deployed in any context that could cause harm to individuals or communities.

\paragraph{Advantages of Using LLMs as Annotators or Generators of CB Data}  
The creation of authentic CB datasets relies heavily on human annotation, which not only strains resources but also exposes annotators to harmful content, raising serious ethical and psychological concerns. Leveraging LLMs as annotators or generators of CB data mitigates these harms by reducing direct human exposure to toxic material. LLM-based generation and labeling can accelerate dataset creation, maintain consistent annotation quality, and provide controlled, ethically safer alternatives for research purposes.

\paragraph{Best Practices for Responsible Use}  
To ensure ethical utilization of synthetic cyberbullying datasets, researchers should adopt best practices including: thorough inspection and filtering of generated content, monitoring for model-specific biases, limiting access to authorized personnel, and using synthetic data solely for research, detection, and moderation purposes. By following these guidelines, the community can leverage synthetic datasets to advance NLP models responsibly while minimizing potential harm or misuse.


\section{Acknowledgements}
This research is supported by the
Disruptive
Technologies Innovation Fund (DTIF) under the
project ``Cilter: Protecting Children Online'' Grant
No.\ DT~2021~0362 from the Department of Enterprise,
Trade and Employment in Ireland and administered by
Enterprise Ireland (EI).
%
%
This research was conducted with the financial support of
Science Foundation Ireland under Grant Agreement
No.\ 13/RC/2106\_P2 at the ADAPT SFI Research Centre at
Dublin City University.
ADAPT, the SFI Research Centre for AI-Driven Digital Content Technology,
is funded by Science Foundation Ireland through the
SFI Research Centres Programme.
For the purpose of Open Access, the author has applied a
CC BY public copyright licence to any Author Accepted Manuscript version
arising from this submission.

\section{Bibliographical References}\label{sec:reference}

\bibliographystyle{lrec2026-natbib}
\bibliography{lrec2026-example}


\appendix
\section{Prompts}
Here we document the prompts we used.
Long lines are wrapped here to the column width and line indention is not shown.
For exact reproduction of white space, please consult the source files.

\subsection{Prompts for Data Generation}
For data generation, we use the prompt template of \citet{kazemi2025synthetic}:


\begin{verbatim}
We are creating sample conversations
to aid in cyberbullying detection.
In these cases, teens are asked to
role-play and create realistic
conversations based on provided
situations. There are 11 students
participating in the conversation.
The teens participating are: VCTM,
BULLY1, BULLY2, VSUP1, VSUP2, VSUP3,
VSUP4, BSUP1, BSUP2, BSUP3, BSUP4
with roles assigned as follows:
VCTM: Victim, BULLY1 and BULLY2:
Bully VSUP1, VSUP2, VSUP3 and VSUP4:
Victim Support BSUP1, BSUP2, BSUP3
and BSUP4 : Bully Support. consider
this case: {Case} and consider this
Type of addressed problem: {Type of
Problem}. Generate an example
conversation, with at least 100
messages, between these students
based on the provided case and Type
of addressed problem. Use profanity
and strong language to create a
realistic dialogue. number each
message in the conversation. Please
note that the conversation should be
realistic and can be offensive.
Please make sure to include
different topics and perspectives in
each conversation
\end{verbatim}

\noindent
We confirm that there is no final full-stop and
that the sentence starting with ``Number each message''
is not capitalized.
Following \citet{kazemi2025synthetic}, the variables
\texttt{\{Case\}} and \texttt{\{Type of Problem\}}
are taken from \citet{sprugnoli-etal-2018-creating}.
There are four cases, producing four prompts for
data generation when applied to the above template.


\paragraph{Case A}
\begin{verbatim}
Your shy male classmate has a great
passion for classical dance. Usually
he does not talk much, but today he
has decided to invite the class to
watch him for his ballet show.
\end{verbatim}

\noindent
Type of Problem:
\begin{verbatim}
Gendered division of sport practices    
\end{verbatim}


\paragraph{Case B}

\begin{verbatim}
Your classmate is very good at
school, but does not have many
friends, due to his/her haughty and
‘teacher’s pet’ attitude. Few days
ago, s/he realised that his/her
classmates brought cigarettes to
school and snitched on them with the
teacher. Now they will be met with a
three days suspension, and they risk
to fail the year.
\end{verbatim}

\noindent
Type of Problem:
\begin{verbatim}
Interference in others’ businesses
\end{verbatim}


\paragraph{Case C}

\begin{verbatim}
Your classmate is very good at
school, and everyone think s/he is
an overachiever. S/He studies a lot
and s/he never goes out. S/He does
not speak much with his/her
classmates, that from time to time
tease him/her for his/her unsocial
life. Things have slightly changed
recently: your classmates mum
convinced teachers to increase the
homework for all the students. A
heedless teacher revealed the
request to the class, and now some
students are very angry at him/her.
\end{verbatim}

\noindent
Type of Problem:

\begin{verbatim}
Lack of independence, parental
intromission.
\end{verbatim}


\paragraph{Case D}

\begin{verbatim}
Your shy classmate is good in all
subjects but in gymnastics. For this
reason, his/her classmates often
tease on him/her when s/he
exercises. Recently, the class has
found out a video on the social
network Musical.ly, where s/he
dances gracelessly, on a 90s song
that no one has never heard before.
\end{verbatim}

\noindent
Type of Problem:
\begin{verbatim}
Web virality
\end{verbatim}

\subsection{Prompt for Labelling}
Note that in this prompt, we use GPT to annotate each message with a binary
\texttt{"is\_harmful"} label, \texttt{CB\_types}, and \texttt{tone\_types},
and to assign a conversation-level \texttt{"final\_status"} label.
However, in this paper, we use only the \texttt{is\_harmful} and \texttt{CB\_types}
annotations and retain the \texttt{tone\_types} and \texttt{final\_status}
labels for future analysis.


\begin{verbatim}
You are an expert annotator of CB
(cyberbullying) conversations. I
will provide you a conversation.
Each turn contains:
- "speaker": the role of the sender
  (e.g., VCTM, BULLY1, VSUP1, etc.)
- "text": the actual message
  content.

Your task is to carefully read the
conversation and annotate each
message according to the following
rules:

Annotation Rules:
1) Read the entire conversation to
   understand context.

2) For each message, output a JSON
   object keyed by its message_id
   (conversation number).
   Each object must include:
   - "is_harmful": "yes" or "no"
   - "CB_types": a list of all
     applicable CB types
     • Must be empty if none apply
     • If "is_harmful" = "yes", must
       contain at least one label
     • Multi-word types must use
       underscores (e.g.,
       Insult_Body_Shame,
       Insult_Discrimination_Sexism)
   - "tone_types": a list of
     applicable tone labels (can be
     empty if none)

3) Tone label definitions:
   - "sarcasm" → when the message
     conveys the opposite of what is
     literally said, often mockingly
     or with ridicule
   - "humorous" → when the message
     contains humor, laughter
     markers (lol, lmao, haha,
     {Face with Tears of Joy}), or
     deliberate joking (even if
     cruel)
   - "hate_speech" → when the
     message mocks, insults, or
     discriminates based on
     protected characteristics
     (color, ethnicity, gender,
     orientation, nationality,
     religion, etc.)

4) Some messages may contain
   multiple harmful parts → assign
   multiple CB_types.

5) After all messages, add a
   top-level field "final_status"
   with one of the following values:
   - "positive" → if the
     conversation ends in support,
     de-escalation, or
     reconciliation
   - "negative" → if it ends without
     support or reconciliation
   - To determine this, consider
     only the last 10 messages.

6) Always return JSON only, with no
   extra explanation text.

CB Types and Examples:

1) Threat_or_blackmail
   Expressions of physical or
   psychological threats, or
   blackmail.
   Examples:
   • [I’ll punch you in the face.]
   • [Do as I asked or I’ll post a
     nude photo of you.]

2) Insult_General
   General insults not covered by
   other categories.
   Examples:
   • [You’re just a dickhead]
   • [Spy!]

3) Insult_Body_Shame
   Criticism based on body shape,
   size, or appearance.
   Examples:
   • [You’re a fatty!]
   • [You have the grace of an
     elephant]

4) Insult_Discrimination_Sexism
   Sexist insults or discrimination
   based on sex, gender, or
   orientation.
   Examples:
   • [You’re too dumb to understand
     this, you’re just a woman.]
   • [What are you, some kind of
     tranny freak]

5) Insult_Discrimination_Racism
   Discrimination based on race,
   skin color, ethnicity,
   nationality, or religion.
   Examples:
   • [You’re a Jewish asshole]
   • [Shitty n****r]
   • [Southerner!]

6) Insult_Attacking_Relatives
   Insults aimed at relatives or
   friends of the victim.
   Examples:
   • [Your mother is so fat she
     wouldn’t fit into the Grand
     Canyon]

7) Curse_or_Exclusion
   Wishing harm, exclusion, or
   rejection.
   Examples:
   • [We don’t want you on the team
     anymore]
   • [Outcast of society]
   • [Shut up!]


8) Defamation
   Revealing embarrassing/defamatory
   info about the victim to a
   public.
   Examples:
   • [I heard that his father lost
     his job and became an
     alcoholic]

9) Sexual_Talk_Harmless
   Consensual or neutral sexual
   talk.
   Examples:
   • [I’d like to kiss you]

10) Sexual_Talk_Harassment
    Unwanted sexual talk or
    harassment.
    Examples:
    • [Send me a nude photo of you!]

11) Encouragement_to_Harassment
    Expressions supporting
    harassment.
    Examples:
    • [Ahah, you’re right!]

12) Other
    Any other harmful utterances not
    listed above.
    Examples:
    • [You’re a shitty 04]

Note:
Harmful messages written by the
victim should be annotated with the
corresponding type of insult.

Example Conversation and Output:

Conversation:
1. BULLY1: "You’re so stupid!"
2. VCTM: "I didn’t mean that,
   sorry."
3. VSUP1: "Hey, let’s calm down."
4. BULLY1: "Wow, you’re such a
   genius {Face with Tears of Joy}"
5. VSUP2: "Enough already. Let’s
   move on"

Expected JSON Output:
{
  "1": {
    "is_harmful": "yes",
    "CB_types": ["Insult_General"],
    "tone_types": []
  },
  "2": {
    "is_harmful": "no",
    "CB_types": [],
    "tone_types": []
  },
  "3": {
    "is_harmful": "no",
    "CB_types": [],
    "tone_types": []
  },
  "4": {
    "is_harmful": "yes",
    "CB_types": ["Insult_General"],
    "tone_types": ["sarcasm",
                   "humorous"]
  },
  "5": {
    "is_harmful": "no",
    "CB_types": [],
    "tone_types": []
  },
  "final_status": "positive"
}
\end{verbatim}

\noindent
The text \texttt{\{Face with Tears of Joy\}}
(appearing twice above)
is replaced with the corresponding Unicode
emoji. The input-output example shown above is part
of the prompt.

\end{document}